\documentclass[runningheads]{llncs}
\usepackage[T1]{fontenc}
\usepackage[backend=biber,style=numeric ,sorting=none]{biblatex}%
\usepackage{amsmath,amsfonts}

\usepackage{amssymb}
\usepackage{algorithmic}
\usepackage{graphicx}
\usepackage{textcomp}
\usepackage{xcolor}
\usepackage{bm}
\usepackage{algorithm}

\usepackage{amsthm}
\usepackage{microtype}
\usepackage{enumitem}
\usepackage[switch]{lineno}
\usepackage{multirow}
\usepackage{booktabs}
\usepackage{color}
\usepackage{makecell}
\usepackage{enumitem}
\usepackage{mathrsfs}
\usepackage{pifont}
\usepackage[section]{placeins}
\usepackage{tabularx}
\usepackage{subfig}
\usepackage{subfloat}
\usepackage{bbding}
\usepackage[misc]{ifsym}
\bibliography{arxiv}

\begin{document}

\title{Towards An Efficient and Effective En Route Travel Time Estimation Framework}
 \titlerunning{Efficient and Effective En Route Travel Time Estimation Framework}
%
%
\author{Zekai Shen\inst{1,2}, Haitao Yuan\inst{3}\textsuperscript{(\Letter)}, Xiaowei Mao\inst{1,2}, Congkang Lv\inst{1,2}, \\ Shengnan Guo\inst{1,2}\textsuperscript{(\Letter)}, Youfang Lin\inst{1,2}, Huaiyu Wan\inst{1,2}}

\authorrunning{Z. Shen et al.}
%
\institute{School of Computer Science and Technology, Beijing Jiaotong University, China \and Beijing Key Laboratory of Traffic Data Analysis and Mining, China \\ \email{\{zkshen, maoxiaowei, congkanglv, guoshn, yflin, hywan\}@bjtu.edu.cn} \and
Nanyang Technological University, Singapore  \\
\email{haitao.yuan@ntu.edu.sg}}
\maketitle 

\setcounter{footnote}{0}

\begin{abstract}
En route travel time estimation (ER-TTE) focuses on predicting the travel time of the remaining route. Existing ER-TTE methods always make re-estimation which significantly hinders real-time performance, especially when faced with the computational demands of simultaneous user requests. This results in delays and reduced responsiveness in ER-TTE services. We propose a general efficient framework U-ERTTE combining an Uncertainty-Guided Decision mechanism (UGD) and Fine-Tuning with Meta-Learning (FTML) to address these challenges. UGD quantifies the uncertainty and provides confidence intervals for the entire route. It selectively re-estimates only when the actual travel time deviates from the predicted confidence intervals, thereby optimizing the efficiency of ER-TTE. To ensure the accuracy of confidence intervals and accurate predictions that need to re-estimate, FTML is employed to train the model, enabling it to learn general driving patterns and specific features to adapt to specific tasks. Extensive experiments on two large-scale real datasets demonstrate that the U-ERTTE framework significantly enhances inference speed and throughput while maintaining high effectiveness. Our code is available at \url{https://github.com/shenzekai/U-ERTTE}

\keywords{En Route Travel Time Estimation  \and Uncertainty Quantification}
\end{abstract}

\section{Introduction}
Travel Time Estimation (TTE) serves as a cornerstone of Intelligent Transportation Systems (ITS)~\cite{yuan2021survey}, enabling the prediction of travel time for specific routes with departure time. This is critical for applications like route planning ~~\cite{jensen2024routing}, navigation ~\cite{yuan2023automatic}, traffic forecasting ~\cite{guo2023self,yuan2024nuhuo}, and online ride-hailing services ~\cite{wang2018learning}. Most methods, known as PRe-route TTE (PR-TTE), focus on predicting the total travel time before departure. However, many real-world scenarios require en route TTE (ER-TTE), which provides real-time estimates while driving. Although some studies ~\cite{fang2021ssml,fan2022metaer} have developed ER-TTE models that integrate traveled route data and dynamic features, they often overlook the need for real-time efficiency, limiting practical application. Thus, developing an efficient and effective ER-TTE framework faces several challenges.

\begin{figure}[!t]
    \centering
    \includegraphics[width=0.9\textwidth]{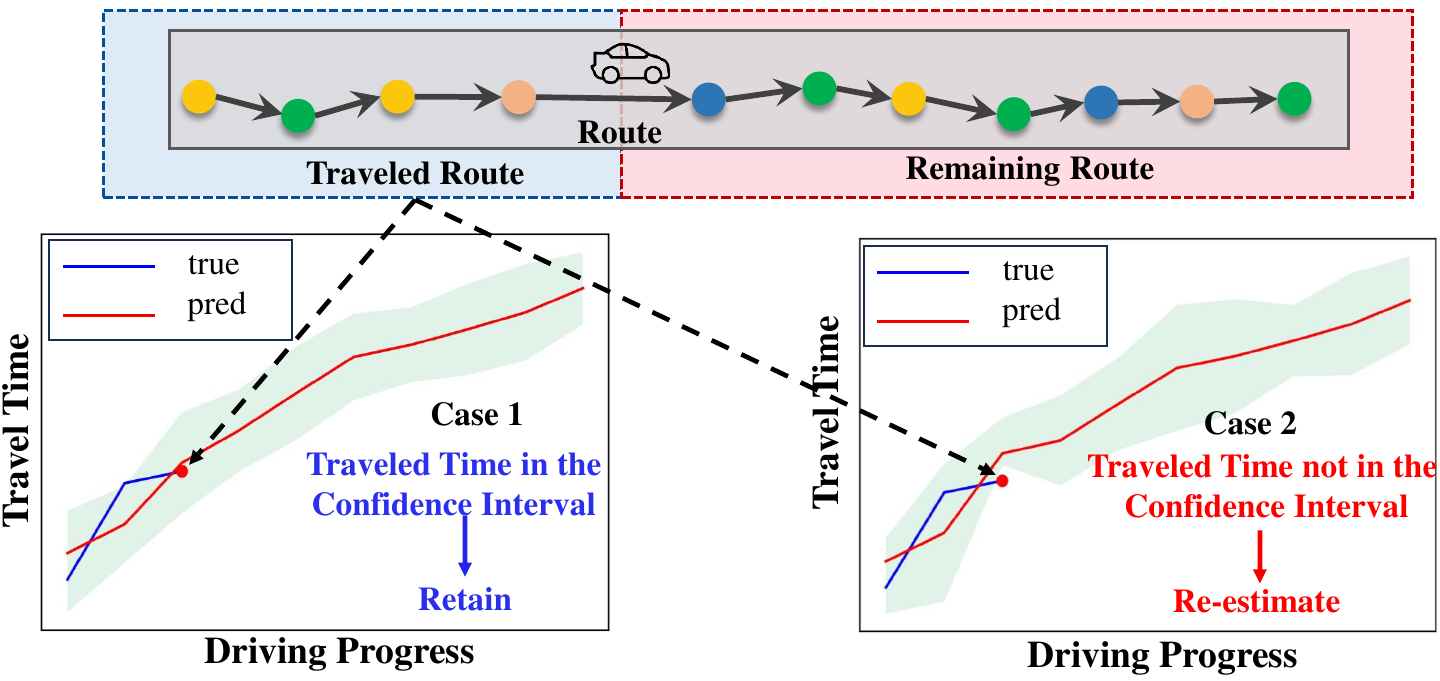}
    \caption{Uncertainty quantification in ER-TTE using confidence intervals to determine when re-estimation is necessary.}
    \label{fig:UQDecision}
\end{figure}
\textit{ \underline{C1:} How to reduce the number of invoking estimation models without loss of accuracy?} To provide timely navigation updates, ER-TTE must refresh remaining travel time estimates during the route. However, frequent model invocations cause delays due to inference time, and high concurrent user requests further strain system capacity. For instance, Didi reported an average of 33.01 million daily orders in China in Q2 2024\footnote{\url{https://ir.didiglobal.com/news-and-events/}}, illustrating the demand for simultaneous ER-TTE updates. Limited server processing power may lead to delays for some users. Therefore, minimizing unnecessary model calls while maintaining accuracy is essential to improving efficiency and throughput.

\textit{ \underline{C2:} How to ensure the robustness of the TTE model throughout the route?} It’s essential to balance general driving patterns with route-specific conditions. The model should learn common travel patterns across routes while quickly adapting to individual driving styles, such as unique acceleration or braking habits. Additionally, it must promptly adjust to unexpected conditions like traffic congestion or accidents to deliver accurate, real-time predictions. However, existing approaches ~\cite{fang2021ssml, fan2022metaer} mainly emphasize capturing user driving preferences. Although useful, these models often lack the flexibility to adapt to dynamically changing traffic situations based on real-time information. For example, during traffic congestions, the models may underestimate delays, which lowers overall prediction accuracy.

In this paper, we introduce an efficient and effective framework U-ERTTE to tackle the key challenges in ER-TTE, featuring an innovative Uncertainty-Guided Decision (UGD) mechanism and Fine-Tuning with Meta-Learning (FTML). The UGD mechanism optimizes requests handling by strategically reusing previous estimates, thus minimizing redundant computations. As shown in Figure~\ref{fig:UQDecision}, the system first quantifies uncertainty across the entire route and stores confidence intervals. During travel, it monitors actual travel time. If travel time remains within the confidence intervals, the estimate is retained, avoiding unnecessary re-estimations. The ER-TTE is then derived by subtracting elapsed time from the initial estimate. Only when significant deviations occur, indicating changed traffic conditions, does the system trigger re-estimation. This UGD mechanism reduces model invocation frequency, addressing the challenge of high request volumes (i.e., \textbf{addressing C1}). Additionally, FTML is introduced to enhance prediction accuracy, using MAML ~\cite{finn2017model} to pre-train on general driving patterns and then fine-tune for specific tasks, allowing rapid adaptation to route dynamics while preserving generalizability (i.e., \textbf{addressing C2}). Therefore, the framework U-ERTTE significantly boost both the effectiveness and efficiency of ER-TTE.

In summary, the contributions can be summarized as follows:

\begin{enumerate}[leftmargin=*]
    \item To our best knowledge, this study is the first to consider the efficiency of ER-TTE and achieve an efficient and effective ER-TTE framework.
    \item We propose an Uncertainty-Guided Decision mechanism (UGD) that determines adaptively whether ER-TTE needs to be re-estimated, offering a novel approach that can be integrated with existing TTE backbone models. 
    \item We develop a new training strategy called Fine-Tuning with Meta-Learning (FTML) that enhances both generalizability and adaptability.
    \item Experiments on two large-scale real-world datasets verify the efficiency and effectiveness of the framework. Our approach significantly improves inference speed and throughput while maintaining high performance.
\end{enumerate}
\section{RELATED WORK}
\subsection{Travel Time Estimation}

Travel time estimation (TTE) can be divided into OD-based TTE, which relies on origin, destination, and departure time~\cite{wang2014travel,lin2023origin,mao2023gmdnet,mao2023drl4route,mao2024dutytte}, and route-based TTE, which incorporates detailed route information. Early methods based on regression or decomposition offered limited accuracy~\cite{chen2024deep}, whereas deep learning techniques have recently achieved significant improvements. RNN-based models capture sequential dependencies in trajectories~\cite{wang2018learning,mao2021estimated}, while graph neural networks and attention mechanisms more effectively integrate spatio-temporal features~\cite{derrow2021eta,fang2020constgat}. Additionally, Transformer-based approaches and variational encoders have been employed to model complex dependencies and uncertainties~\cite{yuan2022route,chen2022interpreting,li2019learning}. In the domain of early remaining travel time estimation (ER-TTE), meta-learning strategies have enhanced adaptability~\cite{fang2021ssml,fan2022metaer}, although they tend to focus on specific locations rather than continuous route progression.

\subsection{Uncertainty Quantification}
Uncertainty quantification methods, widely applied to real-world problems ~\cite{gawlikowski2023survey}, fall into two categories: Bayesian and non-Bayesian approaches. Bayesian methods face the difficulty of directly calculating the posterior distribution, necessitating the use of approximate methods such as variational inference and Markov chain Monte Carlo (MCMC). Bayesian neural networks (BNNs) ~\cite{wang2016towards} utilize approximate Bayesian inference to enhance the efficiency of their inferential processes. Monte Carlo dropout (MC Dropout) ~\cite{gal2016dropout} assumes that the parameters of each layer of the neural network follow a Bernoulli distribution, thereby controlling the drop probability of hidden layer neurons. Non-Bayesian methods, typically frequentist in nature, offer greater flexibility through techniques like quantile regression ~\cite{koenker1978regression} and MIS regression ~\cite{wu2021quantifying}, which embed uncertainty directly into loss functions.

\section{PRELIMINARY}
\begin{figure}[!t]
    \centering
\includegraphics[width=0.9\textwidth]{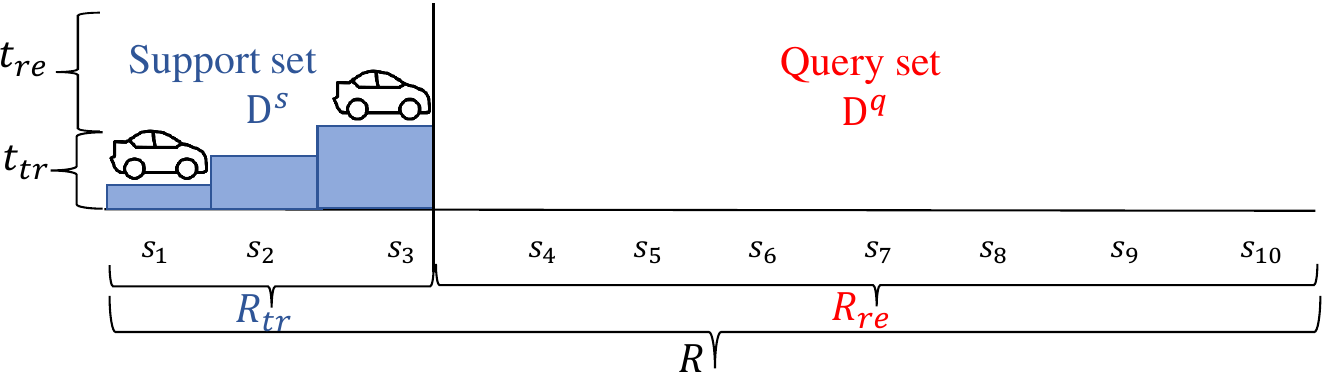}
    \caption{An explanation of Meta-Learning for ER-TTE}
\label{fig:ER-TTE}
\end{figure}
\subsection{En Route Travel Time Estimation}

Given a driving route $ R = [s_1, s_2, \ldots, s_n] $ composed of $ n $ road segments, and considering current segment $ s_m $ ($ 1\le m < n $) at time $ t $, the route can be segmented into two parts: traveled route $ R_{tr} = [s_1, s_2, \ldots, s_m] $ and remaining route $ R_{re} = [s_{m+1}, s_{m+2}, \ldots, s_n] $. 
Specifically, the travel time for the traveled portion of route $R_{tr}$ is known and denoted as $ y_{tr}$.
The objective of the en route travel time estimation (ER-TTE) task is to predict the travel time  
for the remaining portion of  route $ R_{re} $. This task can be formally described by function:
\[
\hat{y}_{re} = f_{\theta}(t, y_{tr}, R_{tr}, R_{re}),
\]
where $ f_{\theta} $ denotes the ER-TTE model, $ \theta $ represents the model parameters, $ t $ is the current time, and $ \hat{y}_{re} $ is the estimated travel time for $ R_{re} $.


\subsection{Meta-Learning in ER-TTE}
Meta-learning, commonly called “learning to learn”, is designed to train models capable of rapidly adapting to new tasks with limited new data. This approach enables a meta-learner to guide the model in effectively learning from diverse tasks. In the context of the ER-TTE problem, meta-learning is employed to discern general patterns of the entire routes, allowing the model to tailor its predictions to the specific nuances of ER-TTE scenarios. Such as the traveled routes contain driver preferences such as acceleration.

In the meta-learning framework applied to ER-TTE, the dataset is strategically split into a support set and a query set. Following the setting of prior research ~\cite{fang2021ssml,fan2022metaer}, the traveled portion of the route $R_{tr}$ is utilized as the support set $\mathrm{D}^{s}$, while the untraveled segment $R_{re}$ is treated as the query set $\mathrm{D}^{q}$. This division aids the model in leveraging past experiences (support set) to make informed predictions about future segments (query set) of the route. A graphical representation of the support and query sets for a specific route is illustrated in Figure \ref{fig:ER-TTE}. This methodological approach underscores the adaptability and foresight that meta-learning imparts to the ER-TTE model.

\begin{figure}[!t]
    \centering
    \includegraphics[width=0.9\textwidth]{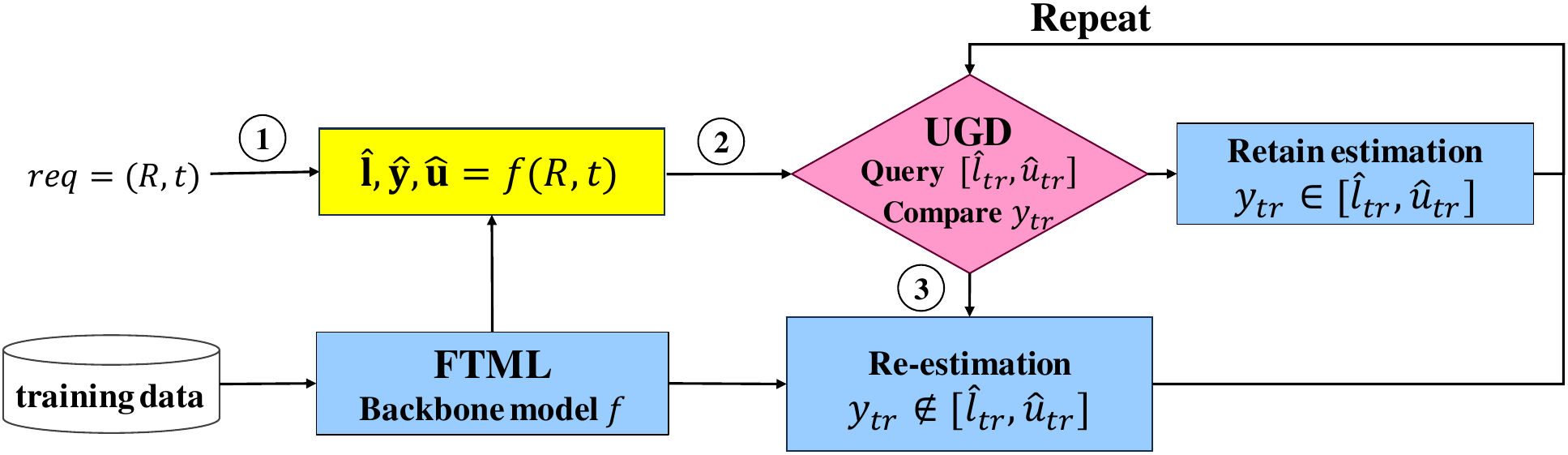}
    \caption{Pipeline of our U-ERTTE framework.}
\label{fig:pipline}
\end{figure}

\section{METHODOLOGY}

\subsection{Overview}
\label{sec:overview}
To enhance the efficiency and accuracy of the ER-TTE process, we propose a simple yet effective framework U-ERTTE as depicted in Figure \ref{fig:pipline}. At first, for a given request $req = (R, t)$  , the model $f$  first estimates the travel time $\hat{\mathbf{y}}$ and the associated confidence intervals for the entire route before departure, utilizing the Fine-Tuning with Meta-Learning (FTML) strategy. The initial confidence intervals, denoted as $ \hat{\mathbf{l}}$ and $\hat{\mathbf{u}} $, encapsulates the predicted travel time from the origin to the destination. Subsequently, during the route, the Uncertainty-Guided Decision (UGD) mechanism is employed to continuously assess the alignment between the actual travel time and the estimated confidence intervals. If the actual travel time $ y_{tr} $ falls within the current confidence interval $ [\hat{l}_{tr}, \hat{u}_{tr}] $, the existing estimation is retained. Conversely, if $ y_{tr} $ deviates from $ [\hat{l}_{tr}, \hat{u}_{tr}] $, a re-estimation is triggered using model $ f $, and this validation cycle is repeated throughout the route to ensure reliable predictions.

\noindent\textit{Remark.} We divide the entire route for $k$ parts. The terms $ \hat{\mathbf{l}}=\{l_1,l_2,...,l_k\}$ and $ \hat{\mathbf{u}}=\{u_1,u_2,...,u_k\} $ represent the lower and upper bounds of the confidence intervals of the entire route, representing the complete range of expected travel times from origin to destination. In contrast, $ [\hat{l}, \hat{u}] $ specifies the confidence interval at a particular point along the route.

\begin{figure}[!t]
\centering
\subfloat[Fine-Tuning with Meta-Learning.\label{fig:FTML}]{
		\includegraphics[width=0.49\textwidth]{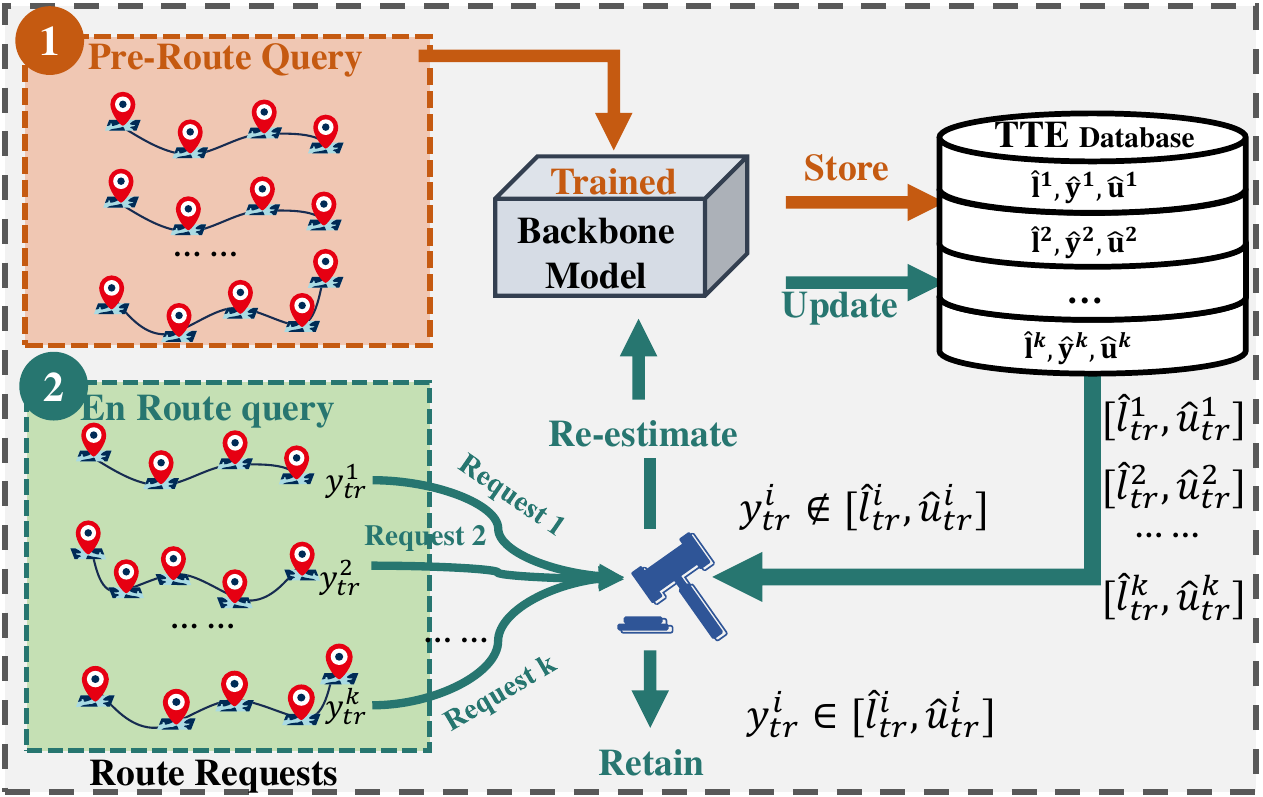}}
\hfill
\subfloat[Uncertainty-Guided Decision.\label{fig:UGD}]{
		\includegraphics[width=0.49\textwidth]{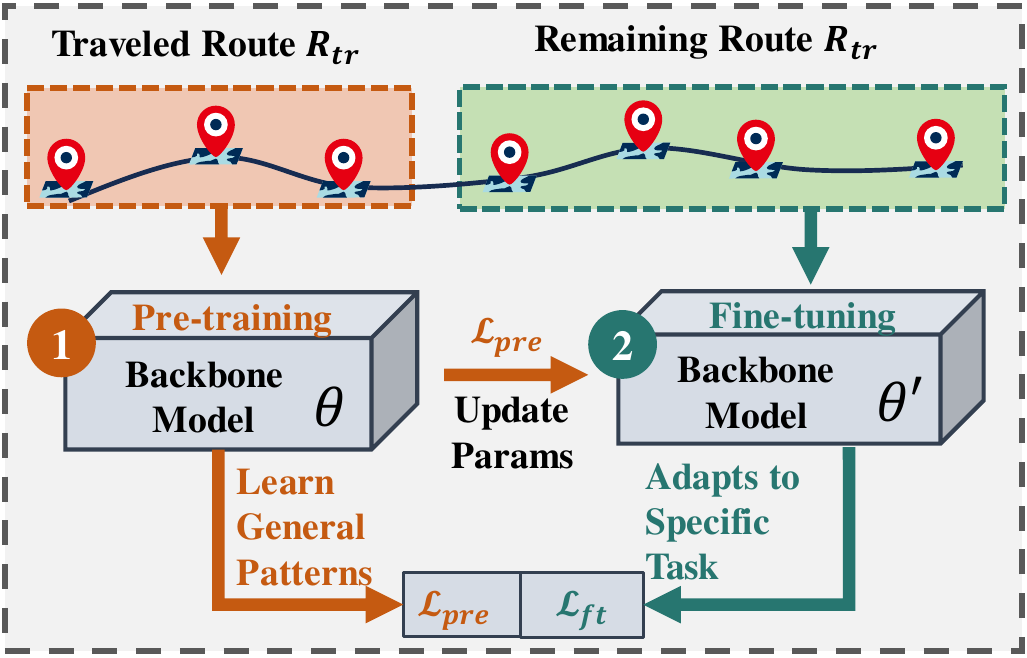}}
\caption{Components of the U-ERTTE Framework}
\label{fig:overall}
\end{figure}

\subsection{Uncertainty-Guided Decision Mechanism}
\label{sec:UGD}
\textbf{Motivation.} To improve the efficiency of ER-TTE and minimize system resource consumption under the large request frequency, it is intuitive to design a decision mechanism to determine whether a re-estimation is needed at the request time $t$ during the driving. Travel time is influenced by unpredictable factors, such as traffic congestion and road accidents, which introduce uncertainty into the predictions. Therefore, quantifying uncertainty allows us to capture the variability in TTE. Accordingly, we propose an Uncertainty-Guided Decision (UGD) Mechanism, quantifying the uncertainty in the route to make decisions. 

\noindent \textbf{Overview.} As illustrated in Figure \ref{fig:UGD}, the Uncertainty-Guided Decision (UGD) process is categorized into two distinct phases: \textit{pre-route query} and \textit{en route query}. Initially, in the pre-route query phase, the trained backbone model is utilized to compute the confidence intervals for each route prior to departure, which are then stored in the TTE database. Subsequently, in the en route query phase, these confidence intervals are employed to verify the accuracy of ongoing predictions. If the actual travel time deviates from the established confidence intervals, indicating the necessity for adjustments, the model is invoked to re-estimate the travel time for the remainder of the route, updating the corresponding confidence intervals in the TTE database accordingly. If the current predictions fall within the confidence intervals, the existing estimations are retained to respond to the en route queries. Notably, the key point is to calculate, store, and update confidence intervals due to that the confidence intervals provide a solid foundation for decision-making.


\begin{algorithm}[htbp]
\caption{Uncertainty-Guided Decision.}
\label{alg:UGD}
\textbf{Require:}\hspace{0.1em} Entire route $R$; traveled routes $\{R_{tr}^1,R_{tr}^2,...,R_{tr}^k\}$; Remained routes $\{R_{re}^1,R_{re}^2,...,R_{re}^k\}$; trained model $f_{\theta}$;
\begin{algorithmic}[1] 
        \STATE  $\mathbf{\hat{l}},\mathbf{\hat{y}},\mathbf{\hat{u}} = f_{\theta}(R,t)$; // \texttt{Pre-route query}
        \STATE  UGD.store(confidence intervals);
        \STATE \textbf{for} $i \longleftarrow 1 ... k $ \textbf{do}: // \texttt{En route query during the route}
        \STATE \quad  UGD.query($y_{tr}^i,R_{tr}^i$) ;
        \STATE  \quad \textbf{if} {${y}_{tr}^i \in [\hat{l}_{tr}^i,\hat{u}_{tr}^i]$}:
        \STATE \quad  \quad $[\hat{l}_{re}^i,\hat{y}_{re}^i, \hat{u}_{re}^i]=[l-\hat{l}_{tr}^i,\hat{y}-\hat{y}_{tr}^i,u-\hat{u}_{tr}^i]$;
        \STATE \quad \textbf{else}: // \texttt{re-estimation}
        \STATE \quad \quad $[\hat{l}_{re}^i,\hat{y}_{re}^i,\hat{u}_{re}^i] = f_{\theta}(t, y_{tr}^i, R_{tr}^i, R_{re}^i);$
        \STATE \quad \quad UGD.update(confidence intervals)
        \STATE \quad \textbf{end for}
\end{algorithmic}
\end{algorithm}

\smallskip
\noindent \textbf{Decision-Making Algorithm.} 
The entire procedure employing the Uncertainty-Guided Decision (UGD) mechanism is outlined in Algorithm \ref{alg:UGD}. Initially, we utilize the model to generate preliminary predictions for the pre-route query, represented as $\mathbf{\hat{y}}$, along with the associated confidence intervals for the entire route, denoted as $\mathbf{\hat{l}}$ (lower bounds) and $\mathbf{\hat{u}}$ (upper bounds). These predictions and confidence intervals are subsequently stored in a database, laying the groundwork for the en route query stage (lines 1-2).
There are $k$ parts in the route which system queries at specific locations.  For each part $i$,the system compares the actual travel time $y_{tr}^i$ with confidence interval $[\hat{l}_{tr}^i, \hat{u}_{tr}^i]$. If the actual travel time resides within the confidence interval, i.e., $y_{tr}^i \in [\hat{l}_{tr}^i, \hat{u}_{tr}^i]$, the prediction is considered accurate, and the original estimate is retained. The ER-TTE is calculated by subtracting the prediction for the traveled route from the initial estimate (lines 5-6). 
Conversely, if the actual travel time falls outside the confidence interval i.e., $y_{tr}^i \notin [\hat{l}_{tr}^i, \hat{u}_{tr}^i]$, this discrepancy suggests significant uncertainty and possible inaccuracies in the prediction, necessitating a re-estimation. In such cases, the system recalculates the travel time using updated route information. (lines 7-9).

Formally, the application of UGD in the en route query to get the remain estimation $[\hat{l}_{re}^i,\hat{y}_{re}^i,\hat{u}_{re}^i]$ can be represented by the following equation:
\begin{equation}
    \label{eq:UGD}
   [\hat{l}_{re}^i,\hat{y}_{re}^i,\hat{u}_{re}^i] = \left\{
        \begin{array}{ll}
            [l-\hat{l}_{tr}^i,\hat{y}-\hat{y}_{tr}^i,u-\hat{u}_{tr}^i] & \text{if } y_{tr} \in [\hat{l}_{tr}, \hat{u}_{tr}], \\
            f_{\theta}\left(t, R_{tr}, R_{re}\right) & \text{if } y_{tr} \notin [\hat{l}_{tr}, \hat{u}_{tr}].
        \end{array}
    \right.
\end{equation}

\noindent \textbf{Efficiency Analysis.} The backbone model complexity is $O(m)$, where $m$ depends on the components of model and input dimensions. When the UGD decides to retain estimation, the need for re-invoking the backbone model is bypassed, thereby reducing the complexity to $O(1)$. Specifically, assuming a confidence level of $p$ (e.g., $p = 0.8$), indicating that there is an 80\% probability that no re-estimation will be needed for a given request during the route, the process predominantly operates at $O(1)$ complexity. As the confidence level increases, the proportion of requests that do not require re-estimation grows, leading to a further reduction in overall time complexity. Consequently, The computational overhead is significantly reduced.

\begin{algorithm}[htbp]
\caption{Fine-Tuning with Meta-Learning.}
\label{alg:FTML}
\textbf{Require}: Entire route $R$; traveled route $R_{tr}$; model parameters $\theta$;  total epoch $\text{N}$; total iteration $n_\text{iter}$.
\begin{algorithmic}[1] 
\STATE \textbf{for} $\text{i}\leftarrow 1 \text{ to } \text{N}$:
   \STATE \quad \textbf{while} $n < n_\text{iter}$:
        \STATE \quad \quad  $\hat{l},\hat{y},\hat{u} = f_{\theta}(R);$ \quad \quad $\hat{l}_{tr},\hat{y}_{tr},\hat{u}_{tr} =f_{\theta}(R_{tr})$.
        \STATE \quad \quad compute the loss function according to Equation \eqref{eq:ptl};
        \STATE \quad \quad Update $\theta'\gets\theta-lr\cdot\nabla_{\theta}\mathcal{L}_{pre}(\theta)$;  // \texttt{pre-train}
        \STATE \quad \quad $\hat{l}_{re},\hat{y}_{re},\hat{u}_{re} = f_{\theta'}(R_{re},R_{tr});$
        \STATE \quad \quad compute the loss function according to Equation \eqref{eq:ftl};
        \STATE \quad \quad Update $\theta \gets\theta-lr\cdot\nabla_{\theta}(\mathcal{L}_{re}(\theta')+\mathcal{L}_{pre}(\theta))$; // \texttt{fine-tune}
   \STATE \quad \textbf{end while}  
\STATE \textbf{Return} $\theta$
\end{algorithmic}
\end{algorithm}
\subsection{Fine-Tuning with Meta-learning}
\label{sec:FTML}

\textbf{Motivation.} To ensure the robustness of TTE models throughout the entire route, on the one hand, the model must effectively learn and generalize driving patterns across various routes to provide a solid foundation for understanding stable driving preferences. On the other hand, the model must be flexible enough to adapt to dynamically changing traffic scenarios, allowing it to adjust predictions based on real-time data. With these goals in mind, we implement Fine-Tuning with Meta-Learning (FTML) to enhance both generalizability and adaptability.

\noindent \textbf{Overview.} As shown in Figure \ref{fig:FTML}, FTML comprises two stages: \textit{pre-training} and \textit{fine-tuning}. In the \textit{pre-training} stage, learns general driving patterns by predicting both the total route arrival time $\hat{y}$ and the traveled time $\hat{y}_{tr}$ at specific locations, along with their confidence intervals. The model parameters are updated to enable the model to learn general driving patterns. In the \textit{fine-tuning} stage, the model further adjusts by predicting the travel time $\hat{y}_{re}$ and the confidence interval for the remaining route, enabling it to adapt to real-time conditions.

\noindent \textbf{Learning Algorithm}. The FTML procedure is depicted in Algorithm \ref{alg:FTML}. In pre-training stage, the model is initialized with effective weights to learn general driving patterns for the entire route and specific features of the traveled route. The focus is on training the model to estimate the travel time $\hat{y}$ and its confidence interval $[\hat{l},\hat{u}]$ for the entire route $R$ as well as the traveled time $\hat{y}_{tr}$ and its confidence interval $[\hat{l}_{tr},\hat{u}_{tr}]$ for traveled route $R_{tr}$ within the support set $\mathrm{D}^s$.
\begin{align}
    \hat{l},\hat{y},\hat{u} &= f_{\theta}(R), \\
    \hat{l}_{tr},\hat{y}_{tr},\hat{u}_{tr} &= f_{\theta}(R_{tr}).
\end{align}
To accurately supervise the predictions and corresponding confidence intervals, we employ \textit{quantile regression} ~\cite{koenker1978regression}, which is a robust and effective method for quantifying uncertainty. Quantile regression is distribution-free and directly optimizes the quantile loss function for providing precise supervision on specific quantile values without requiring any external parameters. In particular, its basic form is illustrated as follows: 
\begin{align}
\label{eq:quantile}
\mathcal{L}^{qua} = &\mathbb{I}_{\hat{l} \geq y} \alpha^{\hat{l}}\left|y-\hat{l}\right|+ \mathbb{I}_{\hat{l}<y}\left(1-\alpha^{\hat{l}}\right)\left|y-\hat{l}\right|+\nonumber\\
&\mathbb{I}_{\hat{y} \geq y}\alpha^{\hat{y}}\left|y-\hat{y}\right|+\mathbb{I}_{\hat{y}<y}\left(1-\alpha^{\hat{y}}\right)\left|y-\hat{y}\right| + \\
&\mathbb{I}_{\hat{u} \geq y} \alpha^{\hat{u}}\left|y-\hat{l}\right|+\mathbb{I}_{\hat{u}<y}\left(1-\alpha^{\hat{u}}\right)\left|y-\hat{l}\right|, \nonumber
\end{align}
where $\alpha$ is the quantile $(0<\alpha<1)$, $\mathbb{I}$ is the indicator function. This loss function is to evaluate the relationship between $\hat{l},\hat{y}, \hat{u}$ and the label $y$, applying different weights based on the target quantile $\alpha$. Specifically, the $\alpha^{\hat{l}}<0.5$, the loss function imposes a greater penalty for cases where the prediction is lower than the label (i.e., lower bound). Conversely, when $\alpha^{\hat{u}}>0.5$, the loss function imposes a greater penalty for cases where the prediction is higher than the label (i.e., upper bound). $\alpha^{\hat{y}}=0.5$, the loss function is for the predication $\hat{y}$.

Building on this, we define a composite loss $L_{pre}$ for the pre-training stage that integrates the quantile loss $\mathcal{L}_{en}$ for the entire route $R$ and the quantile loss $\mathcal{L}_{tr}$ in support set $\mathrm{D}^s$. To further refine the confidence interval, we use Mean Prediction Interval Width (MPIW) ~\cite{pearce2018high} to constrain the confidence interval to ensure it remains tight and reliable throughout the route.
\begin{align}
    \mathcal{L}_{en}(\theta)&=\mathcal{L}^{qua}([\hat{l},\hat{y},\hat{u}],y), \\
    \text{MPIW}&=\hat{u}_{tr}-\hat{l}_{tr},\\
    \mathcal{L}_{tr}(\theta)&=\mathcal{L}^{qua}([\hat{l}_{tr},\hat{y}_{tr},\hat{u}_{tr}],y_{tr})+\text{MPIW}.
\end{align}
The total loss $\mathcal{L}_{pre}$ of the pre-training stage is given by the sum of the two losses (lines 3-4):
\begin{equation}
    \label{eq:ptl}
     \mathcal{L}_{pre}(\theta)=\mathcal{L}_{en}(\theta)+\mathcal{L}_{tr}(\theta).
\end{equation}
Then update the model parameters $\theta$ through the $\mathcal{L}_{pre}$ (line 5):  
\begin{equation}
    \theta'\gets\theta-lr\cdot\nabla_{\theta}\mathcal{L}_{pre}(\theta),
\end{equation}
where $lr$ is the learning rate and $\theta'$ denotes the updated model parameters. By the end of this stage, the model achieves a good weight initialization, enabling it to rapidly adapt to new tasks (i.e.,ER-TTE) in the fine-tuning stage and make accurate predictions. 

During the fine-tuning stage, the inputs for this stage include features of both the already traveled and remaining routes which ensure that the model can adapt to the dynamic traffic conditions of partially traveled routes and then provide accurate ER-TTE. The pre-trained model with parameters $\theta'$ is fine-tuned specifically for prediction $\hat{y}_{re}$ and its confidence interval $[\hat{l}_{re},\hat{u}_{re}]$ for the remaining route in the query set $\mathrm{D}^q$. The fine-tuning quantile loss $\mathcal{L}_{ft}$ for the remaining route is calculated as follows (lines 6-7): 
\begin{align}
    \hat{l}_{re},\hat{y}_{re},\hat{u}_{re} &= f_{\theta'}(R_{re},R_{tr}),\\
    \mathcal{L}_{ft}(\theta') &= \mathcal{L}^{qua} ([\hat{l}_{re},\hat{y}_{re},\hat{u}_{re}],y_{re}).
    \label{eq:ftl}
\end{align}
The model parameters are finally updated based on the two losses (line 8).
\begin{equation}
    \theta \gets\theta-lr\cdot\nabla_{\theta}(\mathcal{L}_{ft}(\theta')+\mathcal{L}_{pre}(\theta)).
\end{equation}

In summary, FTML makes the model generalize from previous tasks (entire and traveled TTE) and swiftly adapt to new tasks (ER-TTE) through a two-stage training process: \textit{pre-training} and \textit{fine-tuning}. During \textit{pre-training}, the model captures general driving patterns for entire routes and the specific features from the traveled route, enabling it to understand the overall dynamics of travel time across different routes. The \textit{fine-tuning} stage allows the model to specialize in the task of ER-TTE, ensuring that it can adapt to the specific conditions of partially traveled routes and provide accurate predictions.

\section{Experiment} \label{Experiment}

\subsection{Experimental Settings}
\noindent \textbf{Datasets and Preprocessing}. We utilize two real-world taxi trajectory datasets collected from the cities of Porto\footnote{\url{https://www.kaggle.com/c/pkdd-15-predict-taxi-service-trajectory-i}},  and Xian\footnote{\url{https://gaia.didichuxing.com/}}.
For both datasets, We removed outlier data (i.e. driving distances that were too short and too long). The processed Porto dataset contains 1,011,761 routes and Xian contains 1,191,125 routes

\noindent \textbf{Metrics}. Similar to existing methods ~\cite{chen2022interpreting}, we use four metrics for performance evaluation, including mean absolute percentage error (MAPE), mean absolute error (MAE), root mean squared error (RMSE), and satisfaction rate (SR). Specifically, SR refers to the proportion of routes with a MAPE less than 10\%, and a higher SR indicates better performance and customer satisfaction. They are defined as follows: $\text{SR}=\frac{1}{N}\sum_{i}^{N}(|\frac{\hat{y_i}-y}{y}|\le 10\%)\times100\%$

\noindent \textbf{Implementation Details}. The Adam optimizer is used with a fixed learning rate of $10^{-3}$ and a weight decay of $10^{-3}$ as a regularization term to prevent overfitting. We use the training set to train the model, select the model with the best MAPE on the validation set, and use the test set to evaluate the performance. All experiments are implemented in Python using the Pytorch toolbox, using an NVIDIA RTX A4000 GPU. The platform runs on an Ubuntu 20.04 operating system. Following  ~\cite{fan2022metaer}, each route is divided into two parts: \textbf{30\%} of the route has already been traveled and \textbf{70\%} remaining route. The quantities $\alpha$ are [0.1,0.5,0.9].

We compare with two strategies that can improve throughput: (1)\textbf{Random}: All samples are not specially processed and are first come, first served. (2)\textbf{Greedy}: Since long routes are often difficult to predict, long-distance routes are predicted first.
We re-estimate samples that could not be effectively filtered under various strategies. For samples that are successfully filtered, the prediction value is calculated as $\hat{y}-y_{tr}$.

\noindent \textbf{Backbone Architectures} We select two groups of backbones including different architecture and integrate them into our framework. As shown in Table \ref{tab:TimeComplexity}, we analyze the time complexity of different backbone models.

\noindent 1. TTE Method:
(1)\textbf{MLPTTE}: A 16-layer multilayer perceptron with ReLU activation function is used. The specific approach is to estimate the travel time of each road segment separately and sum it as the overall travel time estimate of the route.
(2)\textbf{WDR} ~\cite{wang2018learning}: a wide-deep-recurrent architecture is introduced to handle sparse features, dense features, and road segment sequence features respectively.
(3)\textbf{WDR-LC} ~\cite{mao2021estimated}:Enhances WDR by jointly modeling road segments and intersections.
(4)\textbf{ConSTGAT} ~\cite{fang2020constgat}: It is a spatiotemporal graph neural network structure that uses graph attention to capture spatiotemporal correlations and the contextual information of the route.

\noindent 2. ER-TTE Method:
(1)\textbf{SSML} ~\cite{fang2021ssml}: It is the first meta-learning model for ER-TTE, which aims to learn meta-knowledge to quickly adapt to users' driving preferences.
(2)\textbf{MetaER-TTE} ~\cite{fan2022metaer}: A new adaptive meta-learning model is proposed, which generates cluster-aware initialization parameters through soft clustering and uses distribution-aware adaptive learning rate optimization.

\subsection{Overall Effectiveness Comparison}

As shown in Table \ref{tab:effectiveness}, We analyze the effectiveness in two aspects, e.g., Overall comparison and Model fitness.    

\noindent \textbf{Overall Comparison}. We using different backbone models with UGD improves average MAPE, MAE, RMSE and SR by at least 24.3\%, 17.6\%, 12.6\%, and 18\% in on Porto dataset, and 16.7\%, 10.3\%, 7.5\%, and 8.9\% on Xian dataset. This demonstrates the effectiveness of UGD in identifying routes needing re-estimation, and achieving SOTA results. In contrast, the greedy strategy performs worse due to higher prediction errors on longer routes.

\noindent \textbf{Model Fitness}. The attention-based models ConSTGAT, MetaER-TTE, and SSML perform well Because the attention mechanism can integrate information from different segments to obtain accurate predictions. The MLP-based MLPTTE model uses a pure MLP architecture to independently process the features of each road segment and sum them as an estimated value. Although it ignores the association between road segments, it shows advantages in processing independent and complex road segment features.

The RNN-based models WDR and WDR-LC have a slightly poorer performance, which may be because each step of the RNN is calculated based only on the current input and the hidden state of the previous time step. Although RNN can partially integrate historical information, its processing method may not effectively capture the complex associations between different locations, especially when it is necessary to capture the state of the intermediate process of the vehicle's driving process. This causes the RNN-based model to inaccurately estimate travel time, making it difficult to construct an effective confidence interval.

\begin{table*}[!t]

\centering

\caption{Overall performance on Porto and Xian dataset}

\begin{tabularx}{\textwidth}{@{}l|l|XXXX|XXXX@{}}
\toprule

\multirow{2}{*}{Method} & \multirow{2}{*}{Strategy} & \multicolumn{4}{c|}{Porto} & \multicolumn{4}{c}{Xian} \\ 
\cline{3-10}
& & MAPE$\downarrow$  & MAE$\downarrow$  & RMSE$\downarrow$ & SR$\uparrow$ & MAPE$\downarrow$  & MAE$\downarrow$  & RMSE$\downarrow$  & SR$\uparrow$  \\

\hline

\multirow{3}{*}{MLPTTE} & Random & 22.25 & 116.45 & 255.04 & 38.53 & 23.75 & 200.75 & 312.71 & 29.90 \\
& Greedy  & 23.90 & 122.03 & 258.30 & 36.35 & 28.81 & 217.87 & 309.41 & 25.98 \\
& \textbf{UGD}  & \textbf{16.09} & \textbf{91.04} & \textbf{214.19} & \textbf{47.19} & \textbf{18.85} & \textbf{170.83} & \textbf{275.37} & \textbf{35.36} \\

\hline

\multirow{3}{*}{WDR} & Random & 46.67 & 343.86 & 493.88 & 10.61 & 29.78 & 347.44 & 492.96 & 18.56 \\
& Greedy  & 40.22 & 240.24 & 359.06 & 20.86 & 36.13 & 370.64 & 475.88 & 16.21 \\
& \textbf{UGD}  & \textbf{20.08} & \textbf{111.28} & \textbf{236.03} & \textbf{36.71} & \textbf{20.63} & \textbf{193.95} & \textbf{327.10} & \textbf{32.01} \\

\hline

\multirow{3}{*}{WDR-LC} & Random & 32.69 & 234.09 & 384.21 & 24.34 & 32.72 & 381.16 & 530.64 & 16.18 \\
& Greedy  & 44.15 & 287.5 & 417.41 & 14.11 & 36.66 & 363.42 & 476.42 & 19.43 \\
& \textbf{UGD}  & \textbf{19.23} & \textbf{111.27} & \textbf{243.54} & \textbf{40.54} & \textbf{21.88} & \textbf{190.82} & \textbf{294.56} & \textbf{30.34} \\

\hline

\multirow{3}{*}{ConSTGAT} & Random & 23.24 & 121.40 & 257.38 & 37.24 & 25.02 & 209.19 & 319.02 & 28.52 \\
& Greedy  & 26.84 & 129.93 & 252.64 & 27.08 & 25.72 & 215.85 & 323.57 & 26.98 \\
& \textbf{UGD}  & \textbf{15.86} & \textbf{91.76} & \textbf{217.85} & \textbf{48.75} & \textbf{20.83} & \textbf{176.09} & \textbf{269.62} & \textbf{32.82} \\

\hline

\multirow{3}{*}{SSML} & Random & 22.92 & 116.21 & 248.36 & 36.54 & 24.39 & 211.95 & 332.44 & 29.26 \\
& Greedy  & 32.39 & 134.37 & 241.69 & 25.17 & 25.77 & 211.56 & 315.78 & 27.68 \\
& \textbf{UGD}  & \textbf{16.89} & \textbf{90.15} & \textbf{210.03} & \textbf{46.30} & \textbf{19.14} & \textbf{170.10} & \textbf{272.28} & \textbf{35.40} \\

\hline

\multirow{3}{*}{MetaER-TTE} & Random & 25.41 & 129.86 & 264.13 & 31.64 & 24.20 & 200.94 & 304.43 & 29.76 \\
& Greedy  & 32.08 & 145.30 & 254.39 & 19.73 & 27.84 & 256.36 & 379.26 & 21.70 \\
& \textbf{UGD}  & \textbf{16.36} & \textbf{95.96} & \textbf{223.77} & \textbf{44.98} & \textbf{19.98} & \textbf{180.27} & \textbf{281.72} & \textbf{32.42} \\

\bottomrule
\end{tabularx}

\label{tab:effectiveness}

\end{table*}

\subsection{Efficiency Comparison}
\begin{figure*}[!t]
    \centering
    {\includegraphics[width=1\textwidth]{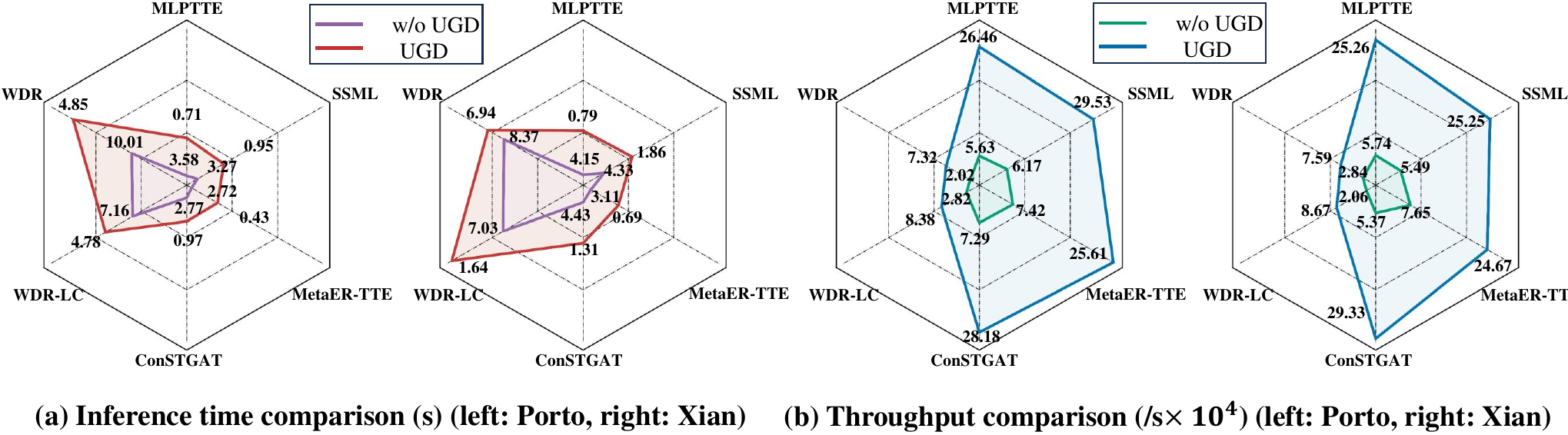}
    \caption{The efficiency comparison of inference time and throughput.}
    \label{fig:efficiency}} 
\end{figure*}
We analyze the efficiency in inference time and throughput(the number of samples that can be processed per second). As shown in Figure \ref{fig:efficiency},  w/o UGD means without UGD which all samples re-estimate. On the Xian dataset, MLPTTE, WDR, WDR-LC, ConSTGAT, and SSML do not require model calls in 77.3\%, 32.1\%, 24.7\%, 77.5\%, and 78.1\% of cases, respectively. On the Porto dataset, these models perform 80.3\%, 74.2\%, 37.9\%, 36.7\%, 73.9\%, and 70.7\%, respectively. We have the following observations: 

(1) The experimental results demonstrate that integrating UGD inference time and throughput by at least 1.49 times and 2.97 times, respectively, on the Porto dataset, and by 1.2 times and 2.67 times on the Xian dataset. 

(2) Three Attention-based models and MLPTTE can filter more samples, they can make greater progress. There are more routes need to re-estimation with two RNN-based methods, resulting in smaller improvements. However, they still improved by 1.2 and 2.67 times in inference time and throughput.

The improvements are attributed to the ability of UGD to minimize unnecessary inference operations, thereby optimizing the use of computational resources and accelerating system response time.

\subsection{Scalability Comparison}
\begin{figure}[!t]
    \centering
    \includegraphics[width=1\textwidth]{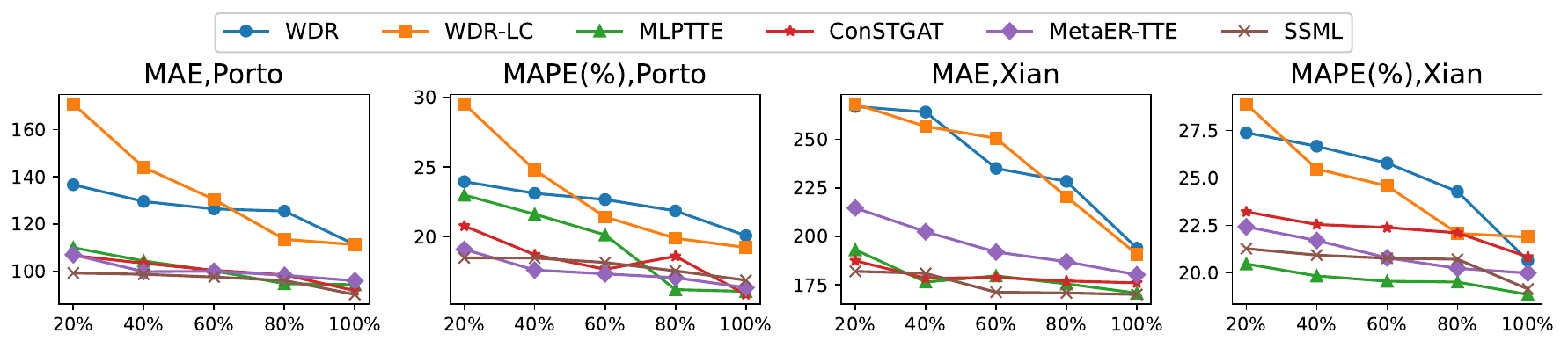}
    \caption{MAPE \& MAE vs. the Scalability.}
\label{fig:scalability}
\end{figure}

To validate the scalability of our framework, we conduct experiments using 20\%, 40\%, 60\%, and 80\% of the training data. As shown in Figure \ref{fig:scalability}, we observe that:

(1)  More data covers a wider range of scenarios, allowing the models to learn more effectively.

(2) Attention-based models achieve good performance even with only 20\% of the data, whereas RNN-based models require larger amounts of data to see significant improvements.

\begin{table*}[!t]
\centering 
\begin{minipage}{0.45\textwidth}
    \centering
    	\caption{Backbone complexity analysis. $n$ refers to the number of segments, $h$ refer to the feature dimensions. }
    	\begin{tabular}{c|cc}
    	\toprule
    		Method & Complexity & Components\\
			\midrule
            MLPTTE     & $O(nh^2)$  & MLP  \\ \hline
            WDR &  $O(nh^2)$ & RNN \\
            WDR-LC  &  $O(nh^2)$ & RNN \\ \hline
            ConSTGAT   & $O(n^2h)$  & Attention  \\
            SSML   & $O(n^2h)$  & Attention  \\
            MetaER-TTE   & $O(n^2h)$  & Attention  \\
    		\bottomrule
    	\end{tabular}
     \label{tab:TimeComplexity}
	\label{datasets} 
\end{minipage}\hfill 
\begin{minipage}{0.45\textwidth}
    \centering
    \caption{Ablation study on the FTML and replace quantile regression to MIS.}
    \begin{tabular}{c|c|cc}
    		\toprule
    		Dataset &  Model & MAPE & MAE  \\ \cline{1-4}
        \multirow{3}{*}{Porto} & w/o FTML &  17.58 & 94.43 \\
        ~ & MIS & 18.46 & 96.88 \\ \cline{2-4}
        ~ & U-ERTTE & \textbf{16.89} & \textbf{90.15} \\ \hline
        \multirow{3}{*}{Xian} & w/o FTML & 20.36  & 176.81   \\
        ~ & MIS & 24.75 & 228.07 \\ \cline{2-4}
        ~ & U-ERTTE & \textbf{19.14} & \textbf{170.10} \\ \hline
    		\end{tabular}
	\label{tab:ablation} 
\end{minipage}
\end{table*}

\subsection{Ablation Study}

To validate our method, we conduct ablation studies on SSML with two modifications: (1) w/o FTML: Removing FTML, and (2) MIS: Replacing the quantile loss with the Mean Interval Score (MIS) loss ~\cite{wu2021quantifying}. As shown in Table \ref{tab:ablation}, removing FTML results in decreased performance, indicating that FTML helps the model capture both general patterns and task-specific features. Moreover, substituting quantile loss with MIS significantly reduces both prediction effectiveness and uncertainty quantification. This demonstrates that quantile loss not only provides accurate confidence intervals to enhance efficiency but also improves overall prediction performance.

\begin{figure}[htbp]
    \centering
    \subfloat[Xian Oct. 2nd 15:39 2322s]{\includegraphics[width=0.44\textwidth]{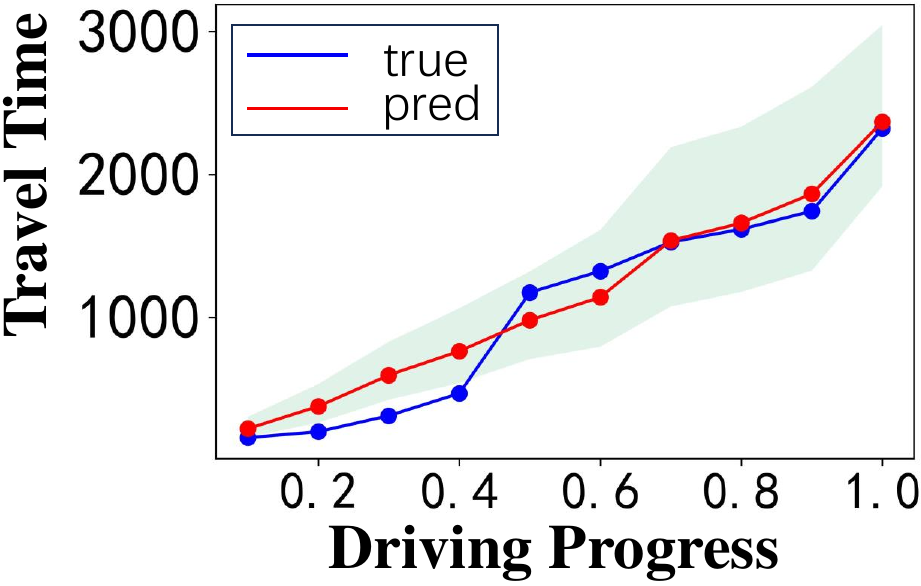}}%
    \label{fig:case1}
    \subfloat[Xian Oct. 14th 23:06 2544s]{\includegraphics[width=0.44\textwidth]{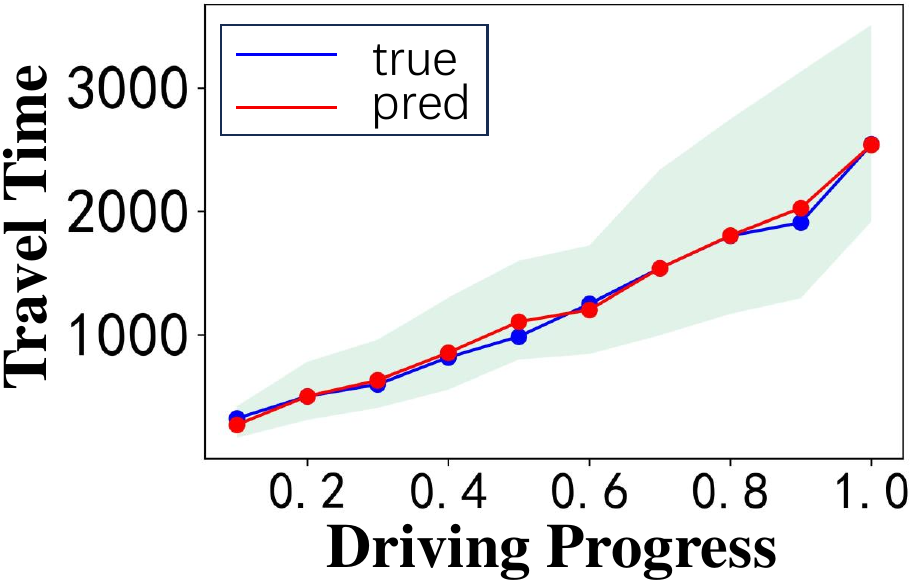}%
    \label{fig:case2}}
    \caption{Visualization of U-ERTTE in ER-TTE process. (The title of each subfigure is labeled in the form of “City, Date, Departure time, and Travel time”.)}
    \label{fig:online}
\end{figure}
\subsection{Online Test}
We conduct an online test simulation on the Xian dataset using a framework with SSML as the backbone model. In this simulation, we select 100 routes that feature both temporal (e.g., weekdays/weekends) and spatial (e.g., short/long distances) characteristics. Each route is divided into 10 equal segments (10\% intervals) to assess online efficiency. Before departure, the model generates confidence intervals for the entire route, and at each 10\% interval (from 10\% to 90\%), an ER-TTE request is made—resulting in 9 queries per route. Although this setup would normally yield 900 requests (9 queries × 100 routes), our framework reduces the number to 334, significantly improving efficiency and throughput.

Figure \ref{fig:online} illustrates the relationship between travel time and route completion percentage for two routes. In the left figure, only 3 initial requests needed re-estimation when actual travel times deviated from predicted confidence intervals, primarily due to holiday traffic in Xian and initial acceleration phases. In contrast, the right figure, which depicts late-night travel, did not require any re-estimations due to stable traffic conditions. These results demonstrate that our framework can efficiently adapt to varying traffic conditions, initiating re-estimations only when necessary, thereby ensuring flexibility and efficiency.

\section{Conclusion}
In this paper, we introduce the U-ERTTE framework, which enhances ER-TTE by integrating UGD and FTML. This framework achieves efficient and effective ER-TTE, an area not previously explored. UGD provides confidence intervals that help the system determine when re-estimation is needed, optimizing computational resource use. FTML improves effectiveness by learning general driving patterns and adapting to specific tasks. This framework advances the state-of-the-art in ER-TTE and offers a practical, scalable solution to enhance system efficiency and performance. In the future, we aim to integrate more diverse real-time data sources, improve scalability, effectiveness, and refine decision mechanisms to handle rare traffic anomalies.
\section*{Acknowledgments}
 This work is supported by the Beijing Natural Science Foundation (Grant No. 4242029).
%
%
%
%
\printbibliography
\end{document}